\documentclass[conference]{IEEEtran}
\IEEEoverridecommandlockouts
\usepackage{cite}
\usepackage{amsmath,amssymb,amsfonts}
\usepackage{algorithmic}
\usepackage{graphicx}
\usepackage{textcomp}
\usepackage{xcolor}
\usepackage{wasysym}
\usepackage{amsmath}

\def\BibTeX{{\rm B\kern-.05em{\sc i\kern-.025em b}\kern-.08em
    T\kern-.1667em\lower.7ex\hbox{E}\kern-.125emX}}
\begin{document}

\title{Structure Analysis of the FRP Rebar Using Computer Vision Techniques\\}

\author{\IEEEauthorblockN{1\textsuperscript{st} Juraj Lagiň}
\IEEEauthorblockA{\textit{Institute of Concrete and Masonry Structures} \\
\textit{Faculty of Civil Engineering} \\
\textit{Brno University of Technology} \\
Brno, Czech Republic \\
lagin.j@fce.vutbr.cz}

\and

\IEEEauthorblockN{2\textsuperscript{nd} Šimon Bilík}
\IEEEauthorblockA{\textit{Department of Control and Instrumentation} \\
\textit{Faculty of Electrical Engineering and Communication} \\
\textit{Brno University of Technology} \\
Brno, Czech Republic \\
bilik@vut.cz}

}

\maketitle

\begin{abstract}
In this paper we present a method to analyze the inner structure of the composite FRP rebar, namely the shift of the real centre of gravity with a respect to the geometrical centre of rebar and changes of cross-sectional characteristics. We propose an automated pipeline based on classical computer vision techniques and on the ratio between the glass fibres and epoxy filament in the analyzed cross-section to compute the shift vector of the real centre of gravity in respect to the geometrical centre together with the cross-section area and its principal moments. We discuss the achieved results over two cross sections in a different portion of the rebar and in the end, we suggest possible direction and improvements for our future work. We also made our code publicly available.
\end{abstract}


\section{Introduction}

    Composite reinforcement is a widely used alternative for reinforcing concrete structures. The FRP (fibre-reinforced-polymer) is known as high tensile strength reinforcement made by \textit{pultrusion} technology. The reinforcement consists of fibres, matrix and surface treatment, which defines material characteristics of rebar. The straight FRP reinforcement is well-known and was described in many books or reviews, such as for example \cite{kniha_FRP}, \cite{FRP_review}. However, for all-composite reinforced structures need to be designed indirect reinforcement as well. The first and oldest design code for bent FRP reinforcement was the JSCE 1997 \cite{JSCE1997}. The next generation of design codes is based on knowledge of this design approach - for example the American ACI 440.1R-15 \cite{ACI4401R15} or Canadian CSA 806-12 \cite{CSA80612} retrieved the approach of determination of ultimate tensile strength for bent FRP reinforcement, which depends on geometrical dimensions of rebar.

   In fact, bent reinforcement is made by bending a straight reinforcement while the manufacturing process, which for thermosetting material leads to to a reduction of the ultimate tensile strength of bent rebar. The subject of this study is to further describe the influence of inner structure bent FRP reinforcement for its ultimate tensile strength. Our study uses computer vision techniques for describing differences in inner structure straight and bent FRP reinforcement.

\section{Related Research}

\begin{figure}[ht]
        \centering
        \includegraphics[width=0.45\textwidth]{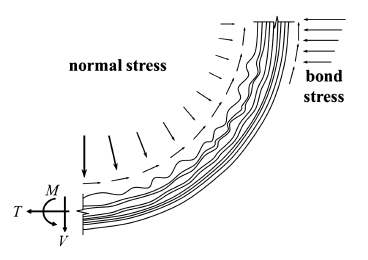}
        \caption{Scheme of kinked kinked fibres along the bent portion of FRP rebar \cite{LeeFRP}}
        \label{fig:Ohyb}
    \end{figure}

    From point of inner structure view, the straight FRP rebars are quite widely described, as can be seen for example in \cite{Robert2010}, \cite{Maligno2008}, \cite{Sevostianov2009}. The research of the inner structure of indirect FRP reinforcement was made just in scale visible by a bare eyed. In the studies \cite{SchockFRP}, \cite{LeeFRP} could be clearly seen the effect of bending while manufacturing process, which leads to kinking fibres on the inner side of the bent portion of rebar and to a flattering of the cross-section of rebar as shown in Fig.~\ref{fig:Ohyb}. This effect causes a reduction of the ultimate tensile strength of bent FRP rebar, and it was already included in design code JSCE 1997 \cite{JSCE1997}.

    Computer vision techniques are today widely used in various applications including traffic \cite{zemvcik2021performance}, industrial inspection \cite{bergmann2018improving} or biology \cite{bilik2021visual}. In this paper, we use classical morphological operations described for example in \cite{sonka2014image}, Otsu method for segmentation threshold estimation \cite{otsu1979threshold} and the region properties analysis over a binary image.

\section{Motivation}

    The main motivation of this research was to describe the differences of the inner structure in the straight and bent portion of FRP rebar. For the purpose of calculating internal forces and stresses in structural engineering, structural mechanics is usually used. The summary is described in many literatures such as \cite{Structural_Mechanics}. Normally, the ultimate tensile strength is defined as a force acting on a rebar area, expressed by the equation:
    
    \begin{equation}
        \label{Rovnice:N/A}
        \sigma_u =\frac {N} {A},
    \end{equation}
    
    where $\mathit{\sigma_u}$ is the ultimate tensile strength, \textit{N} load force, and \textit{A} is the area of the cross-section. As described in the previous capture, the manufacturing process of bent FRP rebar leads to a flattening circle cross-section. The changes cause translocation and shifting of load-bearing fibres to the inner surface of the cross-section. As is the fibre density several times higher, the real centre of gravity of the cross-section is shifted from the geometric centre of gravity. The distance between them is called eccentricity.
    
    When the load force is not acting in the real centre of gravity, it creates an additional bending moment, calculated as:

    \begin{equation}
        \label{Rovnice:Moment}
        M = {N}\cdot{e},
    \end{equation}
        
    where \textit{N} is a load force and \textit{e} is an eccentricity (distance) between the real and geometric centre of gravity.
    
    If the cross-section is loaded by load force and bending moment, the tensile stress in the arbitrary point of the cross-section could be expressed as:

    \begin{equation}
        \label{Rovnice:N/A+M/W}
        \sigma_u =\frac {N} {A} + \frac {M} {I_y}z - \frac {M} {I_z}y,
    \end{equation}
    
    where \textit{M} is the bending moment, $\mathit{I_y}$ and $\mathit{I_z}$ are moment of inertia of cross-section, \textit{z} and \textit{y} are the distance of the considered point from the real centre of gravity of cross-section. The moment of inertia can be calculated:
    
    \begin{equation}
        \label{Rovnice:Iy}
        I_y=\int_A^{} z^2 \,dA,
        \text{  and  } 
        I_z=\int_A^{} y^2 \,dA,
    \end{equation}

    where \textit{A} is an area of cross-section and \textit{z} is the distance of the considered point from the real centre of gravity of the cross-section.
    
    Because the rotation of the cross-section in the image is random, the principal moments of inertia of the cross-section needs to be calculated. The principal moment of inertia also defines a principal axis \textit{1} and \textit{2}. The principal moment of inertia can be calculated as:

    \begin{equation}
        \label{Rovnice:I1,2}
        I_{1,2} =\frac {1} {2}({I_y}+{I_z})\pm\frac {1} {2} \sqrt{({I_y}-{I_z})^2+4{D_{yz}}^2},
    \end{equation}
    
    where $\mathit{D_{yz}}$ is a deviation moment of the cross-section calculated as:

    \begin{equation}
        \label{Rovnice:Dyz}
        D_{yz}=\int_{A}^{} yz \,dA,
    \end{equation}
    
\begin{figure}[ht]
        \centering
        \includegraphics[width=0.45\textwidth]{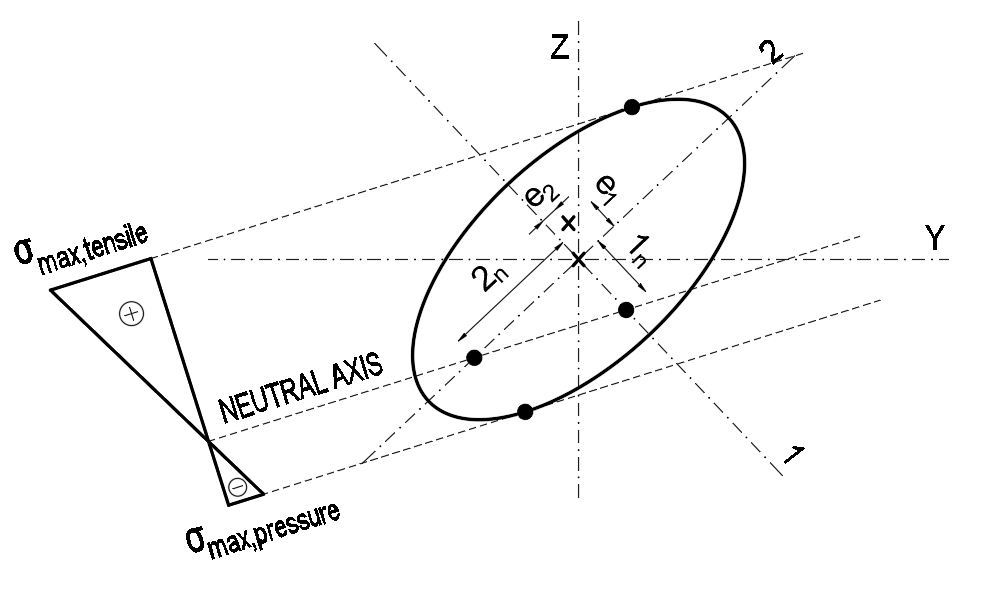}
        \caption{Normal stress according eccentric tensile stress}
        \label{fig:Orientacia}
    \end{figure}

    In a cross-section loaded by a combination of tensile force and bending moment, it is necessary to find the point where the greatest tensile stress is applied. According to structural mechanics, this point can be found based on the neutral axis (see Fig~\ref{fig:Orientacia}), which is defined by the points $\mathit{1_n}$ and $\mathit{2_n}$, calculated as:

    \begin{equation}
        \label{Rovnice:yn,zn}
        1_{n}=\frac {i_{2}^2} {e_1},
        \text{  and  } 
        2_{n}=\frac {i_{1}^2} {e_2},
    \end{equation}
    
    where $\mathit{1_n}$ and $\mathit{2_n}$ are an inertia radius, expressed as:

    \begin{equation}
        \label{Rovnice:iz,iy}
        i_{1}^2=\frac {I_{1}} {A},
        \text{  and  } 
        i_{2}^2=\frac {I_{2}} {A},
    \end{equation}

    In general, it could be assumed that cross-sections with a lower cross-sectional heights have a higher moment of inertia. This leads to higher tensile stress in critical locations of the cross-section in the same magnitude of the load. The aim of our study is to determine the changes in the area of cross-section, translation of the centre of gravity and influence of additional bending moment for the reduction of tensile strength of rebar.

\section{Materials and methods}

    In this section, we describe our data, its collection and the method used to find the shift of the center of the gravity based on the ratio between the fibers and filament masses.

\subsection{Input data description}
    
    The samples for the study were cut from FRP reinforcement, made by E-glass load-bearing fibres, epoxy matrix and sand-coating surface treatment. The reinforcement was made by pultured method, nearly described in \cite{FRP_JL}.

    \begin{figure}[ht]
        \centering
        \includegraphics[width=0.45\textwidth]{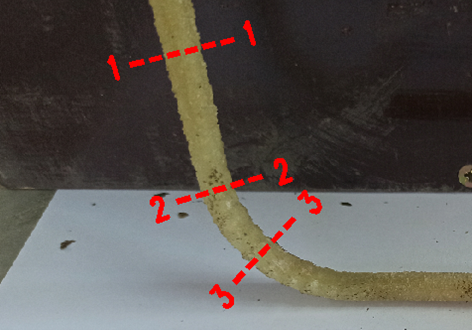}
        \caption{Cut section of experiment samples}
        \label{fig:Rezy}
    \end{figure}  

    The radius of curvature was 30mm and the diameters of rebars \textit{8mm} and \textit{10mm}. Samples were taken from FRP rebar in straight portion (cut n.1) and in a bent portion (cut n.3) as shown in Fig.~\ref{fig:Rezy}.

    \begin{figure}[ht]
        \centering
        \includegraphics[width=0.45\textwidth]{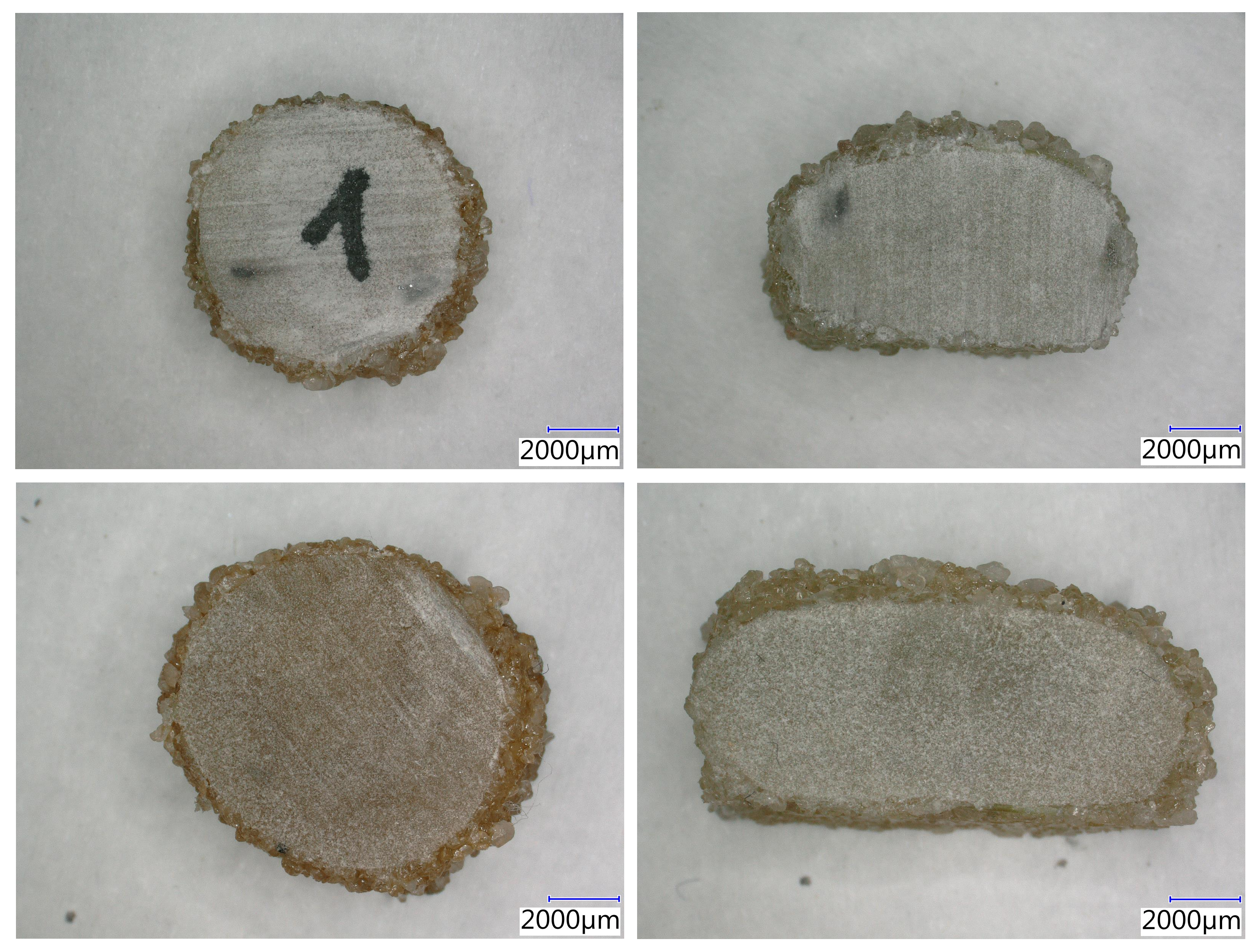}
        \caption{Samples of experiment. From upper left to lower right: \diameter 8\_1-1, \diameter 8\_3-3, \diameter 10\_1-1, \diameter 10\_3-3}
        \label{fig:Vzorky}
    \end{figure}  

\begin{figure}[ht]
        \centering
        \includegraphics[width=0.45\textwidth]{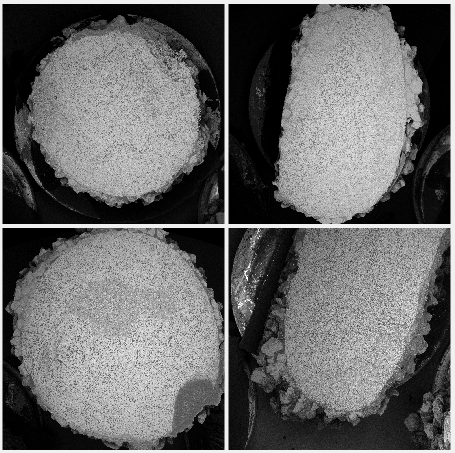}
        \caption{FRP rebar cuts used in our experiment. From upper left to lower right: \diameter 8\_1-1, \diameter 8\_3-3, \diameter 10\_1-1, \diameter 10\_3-3}
        \label{fig:DataSample}
    \end{figure}

    The samples used for our study (see Fig ~\ref{fig:Vzorky})
    were grinded by grinding discs with the roughness of surface \textit{75um}, \textit{35um} and \textit{8um} after cutting. Then they were painted with a gold-plated layer and embedded in the vacuum chamber. Microscopical captures of the samples are shown in Fig.~\ref{fig:DataSample}. The images of samples were obtained by electron microscope at 20 times enlarged.

\subsection{Center of the gravity detection}

    As a first step to detect the center of the gravity, we had to segment the image's ROI, which covers the central section of the cut without the bigger fibers on the side as could be seen in Fig.~\ref{fig:DataSample}. The other areas, which should be removed are the darker grey spots and they are most likely caused by overheating the material while cutting and polishing.

    \begin{figure}[ht]
        \centering
        \includegraphics[width=0.45\textwidth]{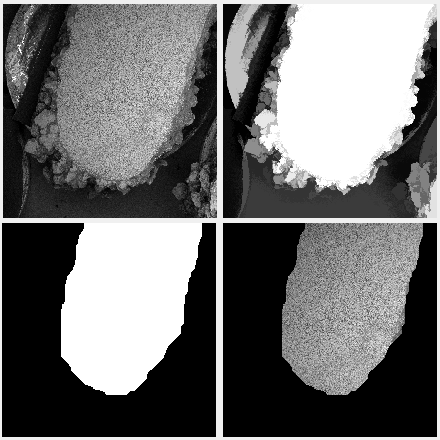}
        \caption{Segmentation steps over the RFB rebar \diameter 10\_3-3 sample. From upper left to lower right: original image, rough segmentation, binary mask, segmented ROI}
        \label{fig:Segmentation}
    \end{figure}

    To achieve such segmentation, we decided to use a cascade of two morphological reconstructions for a rough and fine segmentation. The rough segmentation as a first step is performed over the gray scale image using opening-close reconstruction with the disk kernel of a high diameter. It is followed by multiple Otsu-threshold and it results in a rough mask, which still contains parts of the side fibres.

    The fine segmentation aims to remove the side fibres and the overheated parts on the cross section edges, which is done by unsymmetrical close-opening reconstruction with slightly stronger erosion using a disk kernel. This results into the fine mask and ROI segmentation, as shown in Fig.~\ref{fig:Segmentation}.
    
    In order to compute the real centre of the gravity, we assume that the segmented ROI contains only glass fibres and epoxy filament. To separate them and the background, we again apply multiple Otsu-threshold and we assigned the corresponding values of density and tensile strength to the fibres and epoxy which are together with the ROI used to compute the weighted center.

    \begin{figure}[ht]
        \centering
        \includegraphics[width=0.45\textwidth]{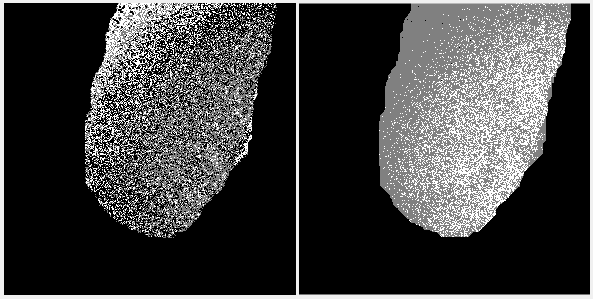}
        \caption{Assigning the weights over the RFB rebar \diameter 10\_3-3 sample. 7x7 local assigning result on the left and global assigning result on the right.}
        \label{fig:Segmentation2}
    \end{figure}

    To assign the corresponding values of densities, we compared the effect of a global and local approach. In the global approach, we considered the brightest pixels as glass fibres and all other pixels under the mask as the epoxy filament. This approach nevertheless did not bring satisfactory results, because the brightness of the cross-section changes with the position. Therefore we used a sliding window of size 7x7 under which we assign the glass fiber density to the pixel with a maximal value and the epoxy density to the pixel of minimal value. Pixels between this range are considered as invalid and we assign them a zero weight. A comparison of the global and local approach is shown in Fig.~\ref{fig:Segmentation2}

    Geometrical centre is computed in a similar way over the binary mask without assigning any weights. In the end, we compute a shift vector between the geometrical and real centre of gravity, which is also displayed in the original image. To compute the distance in \textit{mm} from the distance in \textit{pixels}, we use a scale contained in the header of the microscopic images.

\subsection{Momentum and cross-section points calculation}

    To compute principal moments of inertia $I_1$, $I_2$ together with the intersections of their axis and eccentricity line with the cross-section's mask, we had to first compute the moments $I_z$, $I_y$, deviation moment $\mathit{D_{yz}}$.

    To compute $I_z$, $I_y$ and $\mathit{D_{yz}}$, we iteratively apply morphological erosion with the size of one pixel to the given mask and compute difference between the previous and eroded image. For every pixel of such obtained contour, we compute its contribution to the equations \ref{Rovnice:Iy} and \ref{Rovnice:Dyz} until there were no remaining non-zeros pixels in the difference image.

    Using those values, we compute the principal moments of inertia $I_1$, $I_2$ described with equation \ref{Rovnice:I1,2} and rotation of their axes with the respect to the centre of the gravity axes. Using the $I_1$, $I_2$ axis direction, we compute the intersection points with the given mask and rotate their coordinates with the respect to the $I_1$, $I_2$ axis. The same is applied for the intersection of the eccentricity line in the direction of the centre of the masses shift vector.

    To find the intersection point coordinates, we test if the mask pixel under the given line has value of zero or one. After we find an edge, we save the pixel coordinates as an intersection point in a given direction.

\section{Experimental results}\label{ExpRes}

    As a first part of our experiment, we compare the effect of the sample's shape on the resulting shift vector. Obtained results are shown in Tab.~\ref{tab:Results}, vector size and angle are shown with respect to the Cartesian coordinate system with the beginning in the lower left image corner. The density of the glass fibres was set as 2600 kg/m \textsuperscript{3} and the density of the epoxy filament as 1300 kg/m \textsuperscript{3}.

    \begin{table}[htbp]
    \caption{Resulting shift vectors of the analyzed samples}
    \begin{center}
    \begin{tabular}{|c|c|c|}
    \hline
    \textbf{Sample} & \textbf{Vector size [$mm$]} &	\textbf{Vector angle [°]}\\
    \hline
        \diameter 8\_1-1     &   0.024 & -55.90  \\ 
        \diameter 8\_3-3     &   0.248 & -88.27 \\ 
        \diameter 10\_1-1    &   0.098 & -55.51  \\
        \diameter 10\_3-3    &   0.211 & 56.07 \\ 
    \hline
    \end{tabular}
    \label{tab:Results}
    \end{center}
    \end{table}

\begin{figure}[ht]
        \centering
        \includegraphics[width=0.45\textwidth]{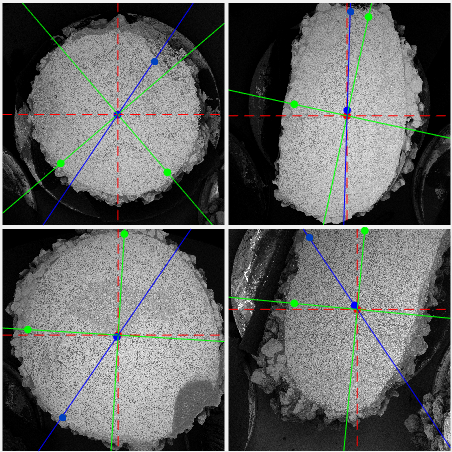}
        \caption{FRP rebar cuts, their centers of the mass (red - original, blue - shifted), axis (red - original, green - principal moments) and cross-section points. From upper left to lower right: \diameter 8\_1-1, \diameter 8\_3-3, \diameter 10\_1-1, \diameter 10\_3-3}
        \label{fig:Shifts}
    \end{figure}

    We could see that the curvature of the rebar significantly affects the resulting shift vector size and angle in the comparison with the non-bended cross-sections - the resulting shift of the elliptic cross-sections \textit{3-3} could be considered as approximately 10x higher than by the circular cross-sections \textit{1-1}. Computations of the geometrical centre of the sample \textit{10\_3-3} seems to be misleading, because the sample was too big to fit in the microscope completely. The resulting center of the mass shifts, axes and intersection points of the analyzed samples are shown in Fig.~\ref{fig:Shifts}.

    \bigskip
    
    Predicted and obtained cross-sectional characteristics compared in the second part of our experiment are shown in Tab~\ref{tab:Results_CSM}: 
   
    \begin{table}[htbp]
    \caption{Resulting cross-section characteristics}
    \begin{center}
    \begin{tabular}{|c|c|c|c|c|}
    \hline
    \textbf{Sample} & {$\mathbf{A_{s,p} [mm^2]}$} &	\textbf{$\mathbf{A_s [mm^2]}$} &	\textbf{$\mathbf{I_1 [mm^4]}$} &	\textbf{$\mathbf{I_2 [mm^4]}$}\\
    \hline
        \diameter 8\_1-1     &   50.27 & 37.45 & 118.60 & 106.31\\ 
        \diameter 8\_3-3     &   50.27 & 38.22 & 208.82 & 66.45\\ 
        \diameter 10\_1-1    &   78.54 & 58.41 & 311.91 & 248.31\\
        \diameter 10\_3-3    &   78.54 & - & - & -\\ 
    \hline
    \end{tabular}
    \label{tab:Results_CSM}
    \end{center}
    \end{table}

    where $\mathit{A_{s,p}}$ is a predicted area of cross-section according to declared diameter by the manufacturer, and $\mathit{A_{s}}$, $\mathit{I_{1}}$, $\mathit{I_{2}}$ are obtained results of the area and principal moment of inertia. As part of the control, comparisons of the computed moment of inertia of a circular cross-section (but also simpler rectangular cross-sections) of the identical cross-sectional area were also processed. The observed deviations are low and comparing the measured and computed results, our proposed method could be considered to be functional.
    
    From the obtained results shown in Tab~\ref{tab:Results_CSM}, we can see a clear decrease in the effective cross-sectional area, which nevertheless may be partly caused by a deviation of our method. The most significant difference can be observed by the moment of inertia, caused by the change in cross-sectional shape.

\begin{figure}[ht]
        \centering
        \includegraphics[width=0.45\textwidth]{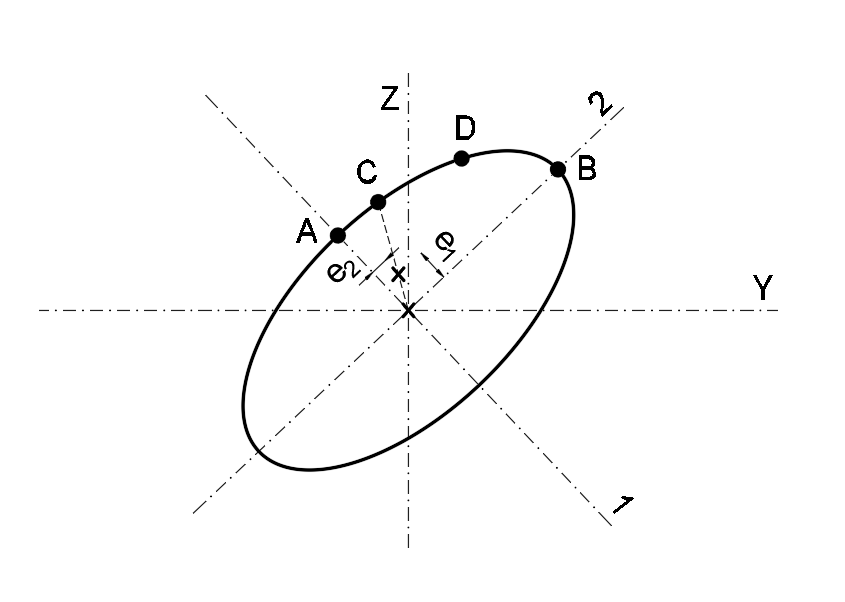}
        \caption{Scheme of points considered for calculation of tensile stress}
        \label{fig:body}
    \end{figure}

    In our previous experiments carried out on a larger number of specimens, we investigated the strength of straight and bent rebars. The resulting tensile strength $\mathit{\sigma_{u,exp}}$ was proved to be related to the declared cross-sectional area. Based on our obtained values shown in Tab~\ref{tab:Results_CSM}, it was possible to apply the experimentally determined average failure force to the 8mm diameter members and then to calculate the stresses in points C, A, E etc. (see Fig~\ref{fig:body}). 
    
    The tensile stresses $\mathit{\sigma_{u,exp}}$ and $\mathit{\sigma_{u,C}}$ were calculated by equation \ref{Rovnice:N/A}. In points A and E is also a bending moment, so the tensile stresses $\mathit{\sigma_{u,A}}$ and $\mathit{\sigma_{u,E}}$ could be calculated by equation \ref{Rovnice:N/A+M/W}. Those calculations were performed only on the samples \textit{\diameter 8} because the sample \textit{\diameter 10\_3-3} was not fully displayed. The obtained results are summarized in Tab~\ref{tab:Results_stresses}:

    \begin{table}[htbp]
    \caption{The tensile stresses in cross-sections}
    \begin{center}
    \begin{tabular}{|c|c|c|c|c|c|}
    \hline
    \textbf{Sample} & \textbf{$\mathit{N_{u,exp}}$} &	\textbf{$\mathit{\sigma_{u,exp}}$} &	\textbf{$\mathit{\sigma_{u,C}}$} &	\textbf{$\mathit{\sigma_{u,A}}$} & \textbf{$\mathit{\sigma_{u,E}}$}\\
        & [kN] & [MPa] & [MPa] & [MPa] & [MPa] \\ 
    \hline
        \diameter 8\_straight     &   68.02 & 1353.28 & 1815.94 & 1856.02 & 1862.16\\ 
        \diameter 8\_bent     &   26.58 & 528.71 & 694.14 & 757.21 & 840.59\\ 

    \hline
    \end{tabular}
    \label{tab:Results_stresses}
    \end{center}
    \end{table}
    
    The results clearly show an increase of the tensile stresses at the centre of gravity of the cross-section compared to the tensile stresses when considering the manufacturer's declared reinforcement diameter. Considering the applied eccentricity, an increase in the tensile stresses at the extreme point \textit{E} was 2.55\% for straight rebar and 21.10\% for bent rebar. If we compare the maximum tensile stress in point E at straight (\textit{\diameter 8\_1-1}) and bent (\textit{\diameter 8\_3-3}) sections of the rebar, we can observe a difference of 54.86\%. This difference indicates that the bending moment is not the only factor causing the reduction of the tensile strength of rebar and as was mentioned in the introduction of the paper, the overall reduction in tensile strength is also caused by the shrinkage of the fibres on the inner side of the bend, which causes that the fibres to lose their effectiveness. However, based on the results obtained in previous experiments, it can be argued that from the total experimentally determined tensile strength reduction of the reinforcement of 60.98\%, approximately 6.12\% can be attributed to the effect of bending moment and change in cross-sectional characteristics. Our code is available at \footnote{https://github.com/boortel/FRP-Rebar-Structure-Analysis}.
          
\section{Conclusion}

    In this study, we analysed microscopic samples of the FRP rebar cuts in order to compute the centre of the gravity shift vector with the respect to the rebar's geometrical centre. To achieve this, we presented a method how to automatically segment the rebar's ROI and how to compute shifted centre of gravity based on the given weights and ratio between the glass fibres and epoxy resin. Based on the obtained results was found the following:
    
    \begin{itemize}
      \item We proved a decrease of the cross-section area based on a comparison of the experimentally obtained value and area computed from a manufacturer-declared diameter.
           
      \item We observed that the centre of gravity translations for samples \textit{\diameter 10\_1-1} and \textit{\diameter 8\_1-1} were insignificant and that they might be caused by a deviation of our method. Sample \textit{\diameter 10\_3-3} shows a translation of the centre of gravity in the direction of the outer surface. However the sample was not fully displayed, so the geometrical centre of gravity is not determined correctly;
      
      For the sample \textit{\diameter 8\_3-3} the translation of the centre of the gravity to the inner surface of the cross-section is several times higher in comparison to \textit{\diameter 8\_1-1}. It suggests, that the fibres of rebar are concentrated on the inner surface of the cross-section;
      
      \item  In the case of straight reinforcement, the additional bending moment caused an increase in tensile stresses of 2.55\%, which seems to be consistent with the measurement deviation mentioned above. However, for the bent rebars there was an increase up to 21.10\% and it can be argued that the effect of the bending moment on the reduction of the tensile strength of the bent reinforcement has been demonstrated;

      \item By our previous experiments, the reduction of tensile strength was determined to be 60.98\%. By comparing the tensile stresses at the point E of the cross-section of straight and bent reinforcement, it can be argued that the effect of bending moment on the overall tensile strength reduction is approximately 6.12\%;

      \item Weight assignment shown at Fig.~\ref{fig:Segmentation2} should be further improved in such way, that we decrease the number of the pixels with the zero weights;

      \item Our experiments were performed on a low amount of samples and for determination of influence for reduction of tensile strength, experimental dataset should be increased;

      \item As a future experiment with a bigger dataset, we suggest to use machine learning methods as semantic segmentation to automatically segment the ROI and to assign the weights. Also, a method should be developed for the automatic calculation of the neutral axis, based on which the point with the maximum tensile stress along the cross section can be determined.
      
    \end{itemize}

\section*{Acknowledgment}

    The completion of this paper was made possible by grant No. FEKT-S-23-8451 - "Research on advanced methods and technologies in cybernetics, robotics, artificial intelligence, automation and measurement" financially supported by the Internal science fund of Brno University of Technology and by the grant "FW01010520 - Development of bent composite reinforcement for environmentally exposed concrete constructions" conducted on Institute of Concrete and Masonry Structures, Faculty of Civil Engineering - Brno University of Technology.

\end{document}